\documentclass[10pt,twocolumn,letterpaper]{article}

\usepackage{cvpr}              % To produce the CAMERA-READY version
\usepackage{graphicx}
\usepackage{amsmath}
\usepackage{amssymb}
\usepackage{multirow}
\usepackage{booktabs}
\usepackage{amsmath,amsfonts,bm,cuted}
\usepackage[ruled,vlined,linesnumbered]{algorithm2e}
\usepackage{graphicx}
\usepackage{tabularx} % in the preamble
\usepackage{amssymb}
\usepackage{multirow}
\usepackage{subfigure}
\usepackage{xcolor}
\usepackage{dsfont}
\usepackage{upgreek}
\usepackage{url}
\def\eqref#1{equation~\ref{#1}}
\def\1{\bm{1}}

\DeclareMathAlphabet{\mathsfit}{\encodingdefault}{\sfdefault}{m}{sl}
\SetMathAlphabet{\mathsfit}{bold}{\encodingdefault}{\sfdefault}{bx}{n}

\newcommand{\softmax}{\mathrm{softmax}}

\DeclareMathOperator*{\argmax}{arg\,max}

\usepackage{times}
\usepackage[accsupp]{axessibility}  % Improves PDF readability for those with disabilities.
\usepackage{pifont}% http://ctan.org/pkg/pifont
\newcommand{\cmark}{\ding{51}}%
\newcommand{\xmark}{\ding{55}}%
\normalsize
\usepackage[pagebackref,breaklinks,colorlinks]{hyperref}

\usepackage[capitalize]{cleveref}
\crefname{section}{Sec.}{Secs.}
\Crefname{section}{Section}{Sections}
\Crefname{table}{Table}{Tables}
\crefname{table}{Tab.}{Tabs.}

\def\cvprPaperID{2} % *** Enter the CVPR Paper ID here
\def\confName{CVPR}
\def\confYear{2022}

\begin{document}

\title{Coarse-to-Fine Reasoning for Visual Question Answering}
\author{Binh X. Nguyen$^{1}$, Tuong Do$^{1}$, Huy Tran$^{1}$, Erman Tjiputra$^{1}$, Quang D. Tran$^{1}$, Anh Nguyen$^{2}$\\
{$^{1}$AIOZ, Singapore}\\
{$^{2}$University of Liverpool, UK}\\
{\tt\small \{binh.xuan.nguyen, tuong.khanh-long.do, huy.tran, erman.tjiputra,quang.tran\}@aioz.io}\\
{\tt\small anh.nguyen@liverpool.ac.uk}}

% \and
\maketitle

\begin{abstract}
Bridging the semantic gap between image and question is an important step to improve the accuracy of the Visual Question Answering (VQA) task. However, most of the existing VQA methods focus on attention mechanisms or visual relations for reasoning the answer, while the features at different semantic levels are not fully utilized. In this paper, we present a new reasoning framework to fill the gap between visual features and semantic clues in the VQA task. Our method first extracts the features and predicates from the image and question. We then propose a new reasoning framework to effectively jointly learn these features and predicates in a coarse-to-fine manner. The intensively experimental results on three large-scale VQA datasets show that our proposed approach achieves superior accuracy comparing with other state-of-the-art methods. Furthermore, our reasoning framework also provides an explainable way to understand the decision of the deep neural network when predicting the answer. Our source codes can be found at: \url{https://github.com/aioz-ai/CFR_VQA}

%To deal with compositional reasoning questions about an image, a Visual Question Answering (VQA) model needs to understand the visual scene, the semantics clues in questions, as well as how the visual and language information interact with each other. Under our observation, besides image features and question features, fine-grained predicates can be extracted to allow a stronger connection when reasoning. However, they are composed of complex semantic information, which is noisy or unnecessary. Thus, prevent the agent from giving out the correct answer. In this paper, we propose a Coarse-to-Fine Reasoning Framework that selectively learns features and predicates of both image and question. The introduced framework allows to reduce the noisy information, then improve the overall accuracy. Moreover, the answer output by our framework can be reasoned explicitly during the prediction progress. The extensive experiments on GQA, VQA 2.0, and Visual7W datasets indicate that our method achieves competitive performance comparing with recent state-of-the-art approaches.

\end{abstract}

% % \vspace{-0.25cm}
\section{Introduction}
\label{sec:intro}

The Visual Question Answering (VQA) task aims to predict the correct answer of a given question such that the answer is consistent with the visual image content.
There are two main variants of VQA, i.e., Free-Form Opened-Ended (FFOE) and Multiple Choice (MC). In FFOE, an answer is a free-form response of a given image-question input pair, while in MC, the answer is chosen from a list of predefined ground-truth. In both cases, extracting meaningful features from the images and questions plays a key role.
Furthermore, mapping the semantic features from the images and questions also strongly affects the results \cite{gordon2018iqa}. Most of the existing solutions for the VQA task rely on visual relations \cite{chen2019routingGraph,cadene2019murel,zhang2020multimodal,yang2020trrnet}, attention mechanisms \cite{teney2017graphvqa,tan2019lxmert, Kim2018BilinearAN}, external knowledge \cite{gu2019externalGraph,li2019perceptual}, or message passing \cite{teney2017graphvqa} to link the visual clue with the associated information in the question.

While both extracting and reasoning the features of the image and question are important for VQA, they are not trivial tasks in practice. Many questions (and answers) are composed of complex semantic information, which can have noise or ambiguous attributes. Current methods focus on utilizing  visual information \cite{chen2019routingGraph,liang2017deepAttGraph,Hudson2019LearningBA,lu2018r,nguyen2021graph,Wang2017FVQAFV,nguyen2020autonomous,narasimhan2018outofboxVQA,Marino2019OKVQAAV} without considering if the supporting information is useful or not \cite{gordon2018iqa}. Besides, many approaches aim to enrich the information extracted from both image and question regardless of the noisy information that may occur \cite{gu2019externalGraph,chen2019routingGraph,gao2020multi,gao2019DFAF}. This leads to the fact that although the image and question features can be extracted by a deep convolutional neural network, they may not be effectively utilized to reason and predict the correct answer.  

To bridge the semantic gap between images and questions in VQA, we introduce a new framework that focuses on reasoning the visual contents in the image and the semantic clues in the question in a coarse-to-fine manner. Our observation is that both image and question's features can be extracted gradually at different fine-grained levels. Therefore, we can map these features in each level to allow a stronger connection when reasoning. Our framework contains effective extractors for extracting meaningful features and predicates from the image and question. Furthermore, the answer outputted by our framework can be reasoned explicitly through the distribution maps during the prediction progress. These maps indicate the necessity of input features or predicates, allow us to understand which information is meaningful for predicting the answer. Our contributions can be summarized as follows:
\begin{itemize}
    \item We propose a simple, yet effective framework to extract meaningful features and predicates from the question and image. The extracted information can be used to explain the decision of the deep network.
    
    \item We introduce a new coarse-to-fine reasoning method to bridge the semantic gap between the question and image when predicting the answer.
    
    \item We conduct intensive experiments to validate our method. Our source code and trained models will be released for further study.
\end{itemize}

% Next, we review the related work in Section~\ref{sec:related_work}. We then describe our CFRF in Section~\ref{sec:input_rep} and Section~\ref{sec:FFOE_vQA}. In Section~\ref{sec:exp}, we present extensive experimental results. Finally, we conclude the paper in Section~\ref{sec:con}.

% We demonstrate the effectiveness and interpretability of our proposed CFRF on  both  FFOE VQA and MC VQA.

% % \vspace{-0.25cm}
\section{Related Work}
\label{sec:related_work}
There are numerous reasoning VQA methods \cite{yi2018neuralreasoning,mao2019neuroSupervision,mascharka2018transparency,teney2021unshuffling,urooj2021found,zheng2021knowledge,zheng2020webly,chen2021meta,hong2021transformation, gao2020multi,do2021multiple,wu2019self,amizadeh2020neuro,luo2021just} that focus on learning the relations between visual regions and words in questions implicitly,  e.g., through message passing \cite{teney2017graphvqa}, pairwise relationship modeling \cite{cadene2019murel}, adversarial learning \cite{li2021adversarial,chi2020collaborative,minh2021deform}, or graph parsing methods defined by inter/intra-class edges \cite{gao2019DFAF}. Other works focus on leveraging external information \cite{gu2019externalGraph} or explicit scene graph \cite{chen2019routingGraph} to extract features from input images. ReGAT \cite{li2019regat} considers both explicit and implicit relations to enrich image representations.
Most of the current VQA works focus on enriching image representation without examining whether the enriched information is necessary for reasoning the answer or not \cite{le2020dynamic}.

Extracting meaningful features from images, questions, and their joint embedding is crucial in the VQA task. For image representation, grid features \cite{jiang2020defense,zhu2016visual7w} or object features \cite{bottom-up2017,do2018affordancenet,tip-trick,nguyen2019object,Ren2015FasterRCNN} are  widely used. For question embedding, Glove \cite{pennington2014glove} and BERT \cite{Devlin2019BERTPO} are used to present words and sentences. Besides, using large-scale pre-training models on image-text pairs is also popular \cite{li2020oscar,chen2020uniter}.
For learning the joint embedding, many approaches use attention mechanisms \cite{teney2017graphvqa,tan2019lxmert,nguyen2019v2cnet,Kim2018BilinearAN, do2019cti,zheng2020cross,zhang2021dmrfnet}. The authors in \cite{Yang2016StackedAN} propose Stacked Attention Networks to localize image regions that are relevant to the question. In~\cite{Kim2018BilinearAN}, the authors propose Bilinear Attention Networks for VQA.
Recently, in \cite{do2019cti}, the authors introduce Compact Trilinear Interaction which simultaneously learns the interaction between images, questions, and answers.

Unlike other approaches that focus on enriching information from image and question, in this work, we consider the interaction among the semantic clues in questions and the visual contents of the image ranging from object-level to fine-grained level. Hence, we apply a simplified fine-grained detector inspired by Faster R-CNN model \cite{Ren2015FasterRCNN} to extract visual features and predicates, rather than leveraging complicated scene graph generators. This setup allows us to achieve competitive results compared with other approaches, while keeping the network at a reasonable computational cost.

% % \vspace{-0.25cm}
\section{Methodology}
\label{sec:input_rep}
\subsection{Overview}

\begin{figure*}[!ht]
    \centering
    \includegraphics[width=\textwidth, keepaspectratio=true]{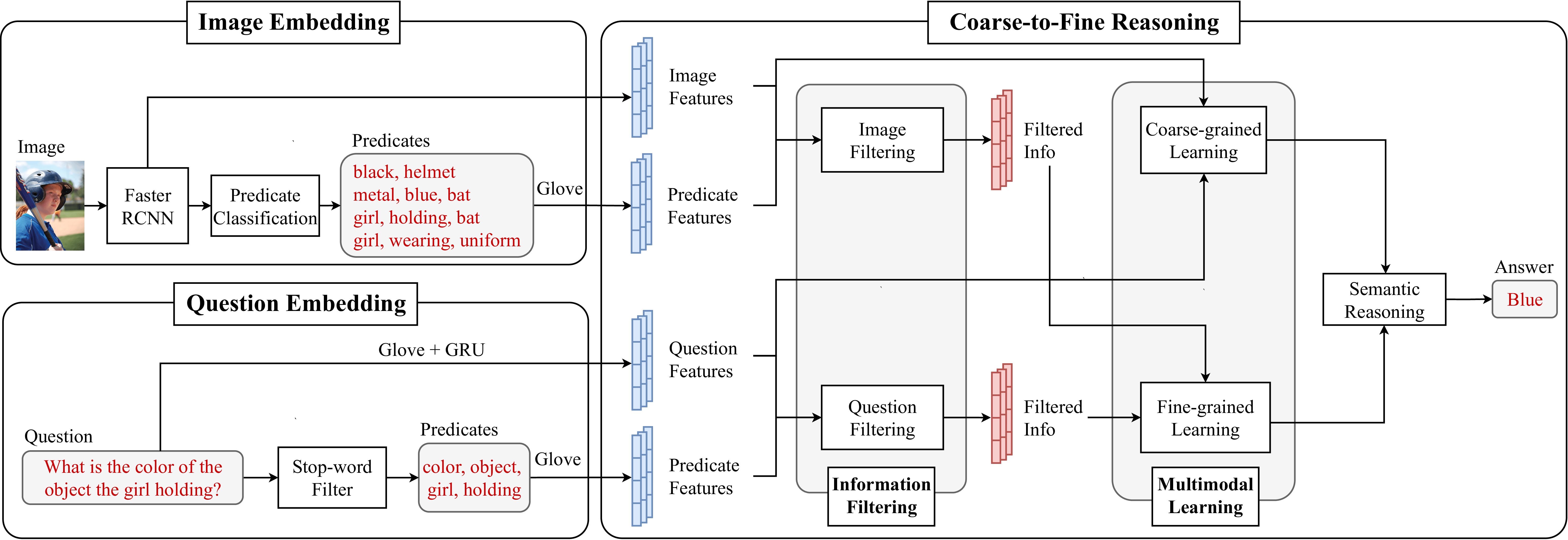}
    \caption{An overview of our framework.}
    \label{fig_overview}
\end{figure*}

Our Coarse-to-Fine Reasoning (CFR) framework takes an image and a question as inputs. The image is passed through the Image Embedding module to extract the region of interest (RoI) features and visual predicates. The question is processed in the Question Embedding module to extract the question features and question predicates. The predicates are keywords about objects, relations, or attributes of the image/question. To effectively map the visual modality and language modality, we jointly learn their features, as well as their predicates in the Coarse-to-Fine Reasoning module. Figure \ref{fig_overview} illustrates an overview of our framework. 

\subsection{Image Embedding}
\label{subsec_image_embedding}

The goal of the Image Embedding module is to extract RoI features and visual predicates from the input image. The RoI features are extracted by a deep object detector to localize all potential regions of interest. The visual predicates are extracted by classifying attributes and relations based on the visual RoI features provided by the object detector.

In practice, as in \cite{Kim2018BilinearAN, tip-trick}, we use the pre-trained Faster R-CNN model \cite{bottom-up2017} to extract visual features for each RoI. Note that the RoI feature is an important visual input for the VQA task. Therefore, we retain the original Faster R-CNN multi-task loss for object detection, then adding two additional Cross-Entropy losses for attribute class predictor and relation class predictor. The extracted objects, as well as their attributes and relations, are then re-arranged to form predicates. Each predicate follows one of three forms: single predicate \texttt{<obj>}; attribute-based predicate \texttt{<attr, obj>}; and relation-based predicate \texttt{<obj1, rel, obj2>}. Following \cite{Kim2018BilinearAN, tip-trick, li2019regat}, we use a pretrained Faster R-CNN model on the Visual Genome dataset \cite{visualgenome} to extract predicates from the images. For each word in each predicate, we apply 300-dim Glove word embedding \cite{pennington2014glove} to extract predicate features.

% With an input image, the proposed Fine-grained Detector $\mathscr{D}$ (See the Fine-grained Detector part in Figure.\ref{fig:CFRF_full}) extracts semantics explicitly, including objects, attributes, and relations between objects.
% As described in Figure \ref{fig:overall}, we apply pretrained Faster R-CNN bottom-up model \cite{bottom-up2017} to extract image features, proposal features, and their positions denoted as Mean Pool Features $M$, RoI Features $R$, and Position features $P$, respectively. 
% About the overall loss function, we retain the original Faster R-CNN multi-task loss and then add two additional Cross-Entropy losses for attribute class predictor and relation class predictor. 

% To predict attributes, each RoI Features $R$ is concatenated with the Mean Pooled Features $M$ and then fed into an additional output layer defining a softmax distribution over each class plus a ``no attribute" class.
% To predict relations, pairs of proposals are generated by considering their position features. Each pair is an ordered $(O, S)$ object tuple, where both $O$ and $S$ are represented by RoI features , denoted as Object Mean Pool Feature $M_O$ and Subject Mean Pool Feature $M_S$, respectively. 
% Then, both mentioned features are concatenated and fed into an additional output layer defining a softmax distribution over each class plus a ``no relation" class. Note that our detector is pre-trained for predicate extraction on Visual Genome dataset \cite{visualgenome}.

\subsection{Question Embedding}

\label{subsec_question_embedding}

The Question Embedding module aims to extract question features and question predicates. To extract question features, following \cite{Kim2018BilinearAN, do2019cti, yu2019mcan}, we apply $600$-dim Glove word embedding \cite{pennington2014glove} accompanied by GRU \cite{2014ChoGRU} to extract the features and learn the dependencies of all words in the question. 

To extract question predicates, we pass the whole question through a stop-word filter. The filter is the combination of two lists. The first list contains words in the NLTK based stop-words \cite{loper2002nltk} list, i.e., words that do not add much meaning in a sentence.  The second list contains words from all the questions that have the frequency of occurrence is less than $10$. Words in the second list are considered as rare words and hard for the model to learn. For each word in each question predicate, we apply 300-dim Glove word embedding \cite{pennington2014glove} to extract the predicate features.

\subsection{Coarse-to-Fine Reasoning}
\label{sec:FFOE_vQA}

Given the image features and predicates ($\mathbf{f}_{\rm i}$,  $\mathbf{p}_{\rm i}$) as well as the question features and predicates ($\mathbf{f}_{\rm q}$,  $\mathbf{p}_{\rm q}$), our goal is to predict an answer $\alpha$ in a list of ground-truth $\mathcal{A}$ using a trainable model $\theta$ as follow:
\begin{equation}
    \hat{\alpha}=\argmax_{\alpha \in \mathcal{A}} \theta\left(\alpha|\mathbf{f}_{\rm i},\mathbf{p}_{\rm i}, \mathbf{f}_{\rm q},  \mathbf{p}_{\rm q}\right)
    \label{eq:general_ffoe}
    % % \vspace{-0.1 cm}
\end{equation}

To effectively map the information of the question to the visual information in the image, the Coarse-to-Fine Reasoning module utilizes three steps: Information Filtering, Multimodal Learning, and Semantic Reasoning. The  Information Filtering aims to filter out unnecessary visual information from the image based on the predicates. The Multimodal Learning module learns the semantic mapping between the question and image at coarse-grained and fine-grained levels. Finally, the Semantic Reasoning module combines the output of the multimodal learning step to predict the answer.

\subsubsection{Information Filtering}
\label{subsec:Instruc_Guiding}
Since the features and predicates of both the question and image are extracted by pretrained models, they may have noise or incorrect information. Therefore, we design the Information Filtering module to filter out unnecessary information. In practice, this module also helps us understand the importance of each RoI for each question. The Information Filtering takes the feature $\mathbf{f} \in \mathds{R}^{\rm n_f \times \rm d_f}$ and the predicate $\mathbf{p} \in \mathds{R}^{\rm n_p \times \rm d_p}$ as input.
% , then outputs the filtered information $\Psi$. 
Both $\mathbf{f}$ and $\mathbf{p}$ have a matrix form; $\rm n_f$, $\rm n_p$ denote the number of instances (e.g., number of RoIs or number of predicates); $\rm d_f$, $\rm d_p$ denote the dimension of each instance. 

To filter out the unnecessary information in the feature $\mathbf{f}$, we consider the predicate $\mathbf{p}$ as the supervision information. Through the interaction mechanism, we compute a weighting map $\hat{\Psi} \in \mathds{R}^{\rm n_f}$ which is then applied to output the filtered information $\Psi \in \mathds{R}^{\rm n_f \times \rm d_{\Psi}}$. $\hat{\Psi}$ is computed as follow:

\begin{equation}
\begin{aligned}
\hat{\Psi}= \softmax\left(\sum^{\rm n_p}_{i=1} \tau _f\left(\mathbf{f}\right) \tau _p\left(\mathbf{p}\right)^T_i\right)
\end{aligned}
\label{eq:instruction_gain}
\end{equation}
where $\tau _f(\cdotp)$ and $\tau _p(\cdotp)$ are learnable linear projection funtions which project $\mathbf{f} \in \mathds{R}^{\rm n_f \times \rm d_f}$ and $\mathbf{p} \in \mathds{R}^{\rm n_p \times \rm d_p}$  into  $\mathbf{f}\sp{\prime} \in \mathds{R}^{\rm n_f \times \rm d_{\Psi}}$  and $\mathbf{p}\sp{\prime} \in \mathds{R}^{\rm n_p \times \rm d_{\Psi}}$ , respectively.

Given the weighting map $\hat{\Psi}$, the filtered information $\Psi$ is calculated by Equation (\ref{eq:instruction_fuse}):

\begin{equation}
\begin{aligned}
\Psi  = \left(\hat{\Psi}\cdot\mathds{1}^T\right) \odot \tau_f\left(\mathbf{f}\right) + \tau_f\left(\mathbf{f}\right)
\end{aligned}
\label{eq:instruction_fuse}
\end{equation}
where $\mathds{1} \in \mathds{R}^{\rm d_{\Psi}}$ is a channel-scaled vector; $\odot$ denotes the Hardamard product.

In practice, the Information Filtering module is applied on both the image features and predicates $(\mathbf{f}_{\rm i}$,  $\mathbf{p}_{\rm i})$, as well as the question features and predicates $(\mathbf{f}_{\rm q}$,  $\mathbf{p}_{\rm q})$ to achieve the filtered information $\Psi_{\rm i}$ and $\Psi_{\rm q}$. Here we use the unified symbol $\Psi$ for simplicity.

%It is worth noting that $\hat{\Psi}$ indicates the importance of each instance when it interacts with predicates. Thus, we can figure out instances that contain unnecessary information. For example, noisy instances in the image can be the RoIs that do not represent any object. In the question, noisy instances can be meaningless words or confusing words.

\begin{figure*}[!ht]
    \centering
    \includegraphics[width=\textwidth, keepaspectratio=true]{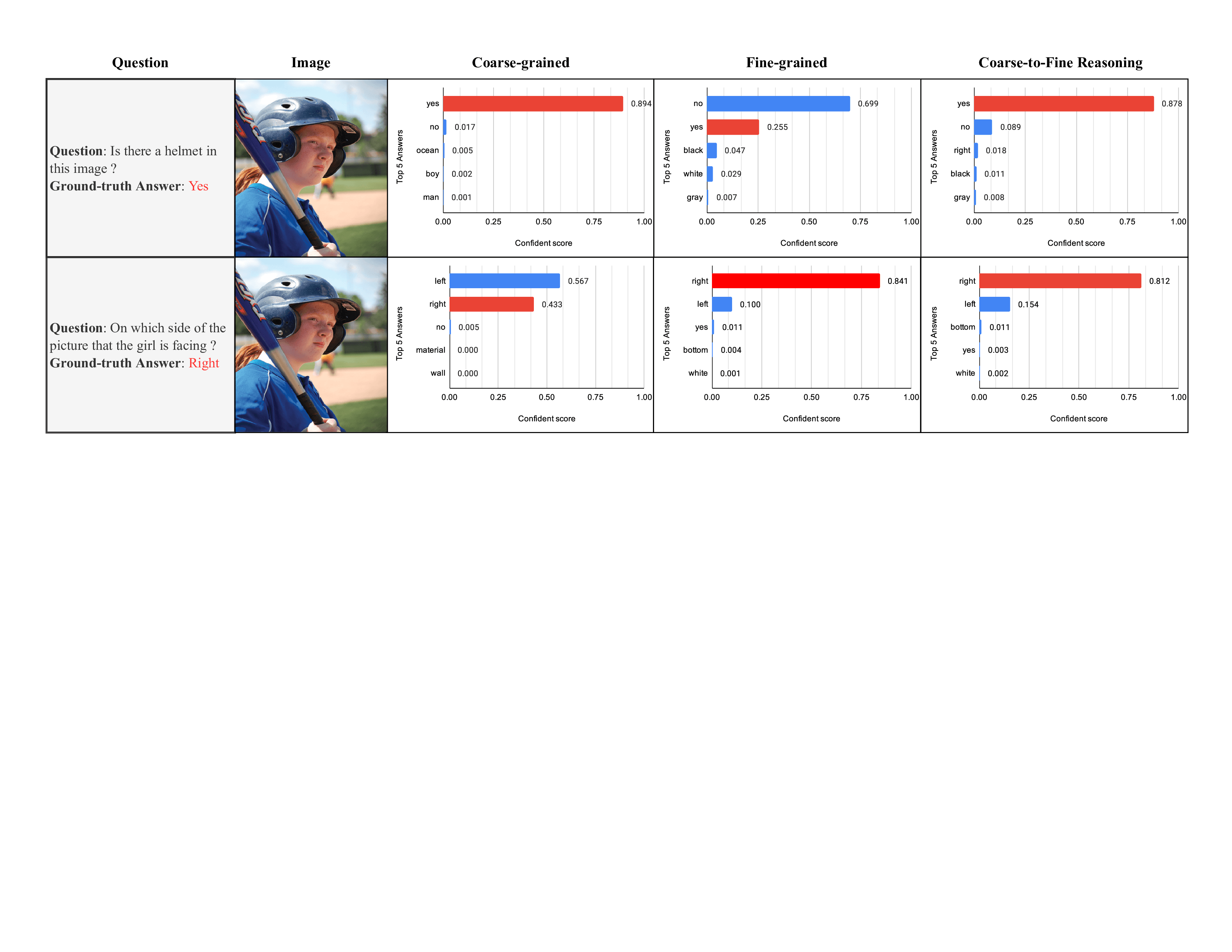}
    \caption{Examples of the predicted confidence scores of the
    Coarse-grained Learning, Fine-grained Learning, and Coarse-to-Fine Reasoning module.}
    \label{fig:sample_Adaptation}
\end{figure*}
\subsubsection{Multimodal Learning}
\label{subsec_multi_integration}

Inspired by the Unitary Attention Mechanism \cite{Kim2018BilinearAN}, we design the Multimodal Learning module to jointly learn the features from the visual and language modalities. Multimodal learning is essential for identifying the correlation between each instance in the image and the question, then identifying which instances in the image are useful for answering the question.

% \textbf{ Differences between BAN and our Multimodal Learning module.} 

% \textbf{Core Differences:} A simplified version of BAN, easier to implement by using only two formulations (4) and (5), instead of (7) formulations in BAN. The results is competitive in comparison with BAN. 

% \textbf{Other differences:} 
% (i) Remove residual learning of attentions. This setup is costly, contains hyper params that may need fine-tuning when dealing with different dataset. (ii) We unify the size of channel-scaled vector in learnable matrices. Hence, further reduce computational cost.

In this module, the features are jointly learned at two levels: coarse-grained and fine-grained. The coarse-grained level learns the interaction between question features and image features, while the fine-grained level learns the interaction between filtered information of the image and question obtained from the Information Filtering step.

%The module then outputs the joint representation $j$ which is further used for answer classification. The bilinear attention mechanism \cite{Kim2018BilinearAN} is leveraged due to its extensive experiments conducted for baseline and benchmarking \cite{do2019cti,kim2020hypergraph,do2020multiple,li2019regat}. 

\textbf{Coarse-grained learning.} The inputs for coarse-grained learning are the image features $\mathbf{f}_{\rm i}$ and question features $\mathbf{f}_{\rm q}$. The output of coarse-grained learning is a joint representation $\mathbf{j}^{\rm cg} \in \mathds{R}^{\rm d_{cg}}$, where $\rm d_{cg}$ is the dimension of the joint representation. Each $k$-th element  of the join representation $\mathbf{j}^{\rm cg}$ is computed as follows:

\begin{equation}
\begin{aligned}
\mathbf{j}_k^{\rm cg} = (\mathbf{f}_{\rm q}\mathbf{M}_{\mathbf{f}_{\rm q}})^T_k\mathbf{A}^{\rm cg}(\mathbf{f}_{\rm i}\mathbf{M}_{\mathbf{f}_{\rm i}})_k
\end{aligned}
\label{eq:bf_fuse_1}
\end{equation}
where $\mathbf{M}_{\mathbf{f}_{\rm q}} \in \mathds{R}^{\rm d_q \times d_{cg}}$ and $\mathbf{M}_{\mathbf{f}_{\rm i}}\in \mathds{R}^{\rm d_i \times d_{cg}}$ are learnable factor matrices%\footnote{Learnable factor matrix: the so-called model weights that are initialized randomly at the beginning of the training phase and change through the learning process.}
; $\rm n_q$, $\rm n_i$ denote the number of instances in question and image;  $\mathbf{A}^{\rm cg} \in \mathds{R}^{\rm n_q \times n_i} $ is the bilinear attention distribution map of the joint representation $\mathbf{j}^{\rm cg}$; $\rm d_q$, $\rm d_i$ denote the dimension of each instance. The subscript $k$ indicates the index of matrix column. $\mathbf{A}^{\rm cg}$ is computed by Equation (\ref{eq:bf_att_1}):

\begin{equation}
\begin{aligned}
\mathbf{A}^{\rm cg} = \softmax\left(\left(\mathbf{f}_{\rm q}\mathbf{M'}_{\mathbf{f}_{\rm q}}\right)\left(\mathbf{f}_{\rm i}\mathbf{M'}_{\mathbf{f}_{\rm i}}\right)^T\right)
\end{aligned}
\label{eq:bf_att_1}
\end{equation}
where $\mathbf{M'}_{\mathbf{f}_{\rm q}}\in \mathds{R}^{\rm d_q \times d_{cg}}$ and $\mathbf{M'}_{\mathbf{f}_{\rm i}}\in \mathds{R}^{\rm d_i \times d_{cg}}$ are learnable factor matrices, and independent of $\mathbf{M}_{\mathbf{f}_{\rm q}}$ and $\mathbf{M}_{\mathbf{f}_{\rm i}}$.

\textbf{Fine-grained learning.} We apply the same process of coarse-grained learning for fine-grained learning. The only difference is the inputs for fine-grained learning are the image filtered information $\Psi_{\rm i}$ and question filtered information $\Psi_{\rm q}$. Similar to Equation \ref{eq:bf_fuse_1} and \ref{eq:bf_att_1}, the fine-grained joint representation is computed as follow:
\begin{equation}
\begin{aligned}
\mathbf{j}_k^{\rm fg} = (\Psi_{\rm q}\mathbf{M}_{\Psi_{\rm q}})^T_k\mathbf{A}^{\rm fg}(\Psi_{\rm i}\mathbf{M}_{\Psi_{\rm i}})_k
\end{aligned}
\label{eq:bf_fuse_2}
\end{equation}
where $\mathbf{A}^{\rm fg}$ is computed as:
\begin{equation}
\begin{aligned}
\mathbf{A}^{\rm fg} = \softmax\left(\left(\Psi_{\rm q}\mathbf{M'}_{\Psi_{\rm q}}\right)\left(\Psi_{\rm i}\mathbf{M'}_{\Psi_{\rm i}}\right)^T\right)
\end{aligned}
\label{eq:bf_att_2}
\end{equation}

% \textbf{Fine-grained learning.} The inputs for fine-grained learning are the image filtered information $\Psi_{\rm i}$ and question filtered information $\Psi_{\rm q}$. The output of fine-grained learning is a joint representation $\mathbb{j} \in \mathds{R}^{\rm d_{fg}}$ where $\rm d_{fg}$ is the intermediate dimension. Each $l$-th element of the join representation $\mathbb{j}$ is computed as follow:

% \begin{equation}
% \begin{aligned}
% \mathbb{j}_l = (\Psi_{\rm q}\mathbf{M}_{\Psi_{\rm q}})^T_k\mathbf{A}_\mathbb{j}(\Psi_{\rm i}\mathbf{M}_{\Psi_{\rm i}})_l
% \end{aligned}
% \label{eq:bf_fuse_2}
% \end{equation}
% where $\mathbf{M}_{\Psi_{\rm q}}$ and $\mathbf{M}_{\Psi_{\rm i}}$ are learnable factor matrices. Note that $\mathbb{j}_l$ is the $l$-th element of $\mathbb{j}$; the subscript $l$ for the matrices indicates the index of column. $\mathbf{A}_\mathbb{j}$ is the bilinear attention distribution map of $\mathbb{j}$ which is computed by Equation (\ref{eq:bf_att_2}):
% \begin{equation}
% \begin{aligned}
% \mathbf{A}_\mathbb{j} = \softmax\left(\left(\Psi_{\rm q}\mathbf{M'}_{\Psi_{\rm q}}\right)\left(\Psi_{\rm i}\mathbf{M'}_{\Psi_{\rm i}}\right)^T\right)
% \end{aligned}
% \label{eq:bf_att_2}
% \end{equation}
% where $\mathbf{M'}_{\Psi_{\rm q}}$ and $\mathbf{M'}_{\Psi_{\rm i}}$ are learnable factor matrices.

\subsubsection{Semantic Reasoning}
\label{subsec:Sec_Adapt}

The goal of Semantic Reasoning is to selectively learn information from both the Coarse-grained and the Fine-grained learning steps using a learnable adaptive weight $\mathbf{W} \in \mathds{R}^{|\mathcal{A}|}$, where $|\mathcal{A}|$ is the number of possible answers. In practice, this module takes $\rm \mathbf{j}^{cg}$ and $\rm \mathbf{j}^{fg}$ as inputs and then outputs the distribution $\rho \in \mathds{R}^{|\mathcal{A}|}$ over candidates of all answers $\mathcal{A}$.

\begin{equation}
\begin{aligned}
&\rho = \softmax \left(\mathbf{W}\tau\left({ \mathbf{j}^{\rm cg}}\right)+{\mathbf{W}'}{\tau'}({\mathbf{ j}^{\rm fg}})\right)\\
&\text{s.t} \sum  \left|\left|\left[{\mathbf{W}_\alpha, \mathbf{W}_\alpha'} \right]\right|\right| = 1, \forall \mathrm{\alpha} \in \mathcal{A}
\end{aligned}
\label{eq:adapt_fuse}
\end{equation}
where $\mathbf{W}$ and $\mathbf{W}'$ are the learnable adaptive weights of coarse-grained learning and fine-grained learning; $\tau(\cdotp)$ and $\tau'(\cdotp)$ are learnable projection functions that project $\mathbf{j}^{\rm cg} \in \mathds{R}^{\rm d_{cg}}$
and 
$\mathbf{j}^{\rm fg} \in \mathds{R}^{\rm d_{fg}}$
into  
$\rho^{\rm cg} \in \mathds{R}^{|\mathcal{A}|}$ 
and 
$\rho^{\rm fg} \in \mathds{R}^{|\mathcal{A}|}$ , respectively. 
To satisfy the constraint in Equation (\ref{eq:adapt_fuse}), we apply the softmax function for each vector $\left[\mathbf{W}_\alpha, \mathbf{W}_\alpha'\right]$; the subscript $\alpha$ indicates the index of an answer in the answer list $\mathcal{A}$.

Through an end-to-end training process, the learned adaptive weights $\mathbf{W}$ identify the contribution of each input information to predict the answer. These weights are expected to robust with noisy information from the question or image at both the coarse-grained and the fine-grained level.

\section{Experiments}
\label{sec:exp}
\subsection{Dataset, baseline and evaluation protocol}
\textbf{Dataset.} We use three popular datasets in our experiments: GQA  \cite{hudson2019gqa}, VQA 2.0 \cite{vqav22016}, and Visual7W  \cite{zhu2016visual7w} . We follow the same split in each dataset for training and testing.

\textbf{Implementation.}
\label{subsec:implement}
We conduct experiments on an NVIDIA TITAN V 12GB GPU. The network is trained with a batch size of $32$ and a learning rate of $0.001$ using Adam optimizer. Following \cite{Kim2018BilinearAN, kim2020hypergraph,tip-trick,Yang2016StackedAN}, we use the Visual Genome \cite{visualgenome} and Glove \cite{pennington2014glove} to extract the image embedding and question embedding. Then we train the whole framework from scratch. The parameters $\rm d_{cg}$ and $\rm d_{fg}$ are empirically set to $768$. The learnable factor matrices $\mathbf{M}$, $\mathbf{W}$ are initialized randomly at the beginning of the training phase and being learned through the training process. It takes approximately $10$, $20$, and $35$ hours to train our network on Visual7W, VQA2.0, and GQA dataset, respectively. 

\textbf{Baselines.} We compare our results with various recent methods in VQA. These methods can be categorized into three groups: joint learning mechanisms: BAN \cite{Kim2018BilinearAN}, Pythia \cite{Jiang2018PythiaVT}, DFAF \cite{gao2019DFAF}, fPMC \cite{hu2018learningfPMC}, STL \cite{wang2018structuredSTL}, CTI \cite{do2019cti}, and MCAN \cite{yu2019mcan}; reasoning-based methods: Murel \cite{cadene2019murel}, ReGAT \cite{li2019regat}, MMN \cite{chen2021meta}, NMS \cite{Hudson2019LearningBA}, and HAN \cite{kim2020hypergraph}; and large-scale visual-language modeling: LXMERT \cite{tan2019lxmert}, OSCAR \cite{li2020oscar}, and UNITER \cite{chen2020uniter}. 

\textbf{Evaluation Metrics.} 
As the standard practice, we use the accuracy metric (\textit{Acc})~\cite{VQA} to evaluate the free-form opened ended dataset (GQA and VQA 2.0), and  \textit{Acc-MC}~\cite{zhu2016visual7w} to evaluate the multiple-choice dataset (Visual7W).

\begin{table}[!t]
\begin{center}
%\footnotesize
\small
% \vspace{+0.15 cm}
\setlength{\tabcolsep}{0.3 em} % for the horizontal padding
{\renewcommand{\arraystretch}{1.2}% for the vertical padding
\begin{tabular}{c|c|c|c|c|c|c}
\hline
\multirow{4}{*}{\textbf{Method}}                    & \multicolumn{6}{c}{\textbf{Dataset}} \\ \cline{2-7} 
& \multicolumn{2}{c|}{\textit{\textbf{GQA (Acc)}}} & \multicolumn{2}{c|}{\textit{\textbf{VQA 2.0 (Acc)}}} & \multicolumn{2}{c}{\textit{\textbf{\begin{tabular}[c]{@{}c@{}}Visual7W\\ (Acc-MC)\end{tabular}}}} \\ \cline{2-7} 
& \textbf{val}          & \textbf{tes-dev}         & \textbf{val}            & \textbf{test-dev}          & \textbf{val}               & \textbf{test}              \\ \hline
BAN \cite{Kim2018BilinearAN} & 61.5                  & 55.2                     & 66.0                    & 70.0                       & 65.7                       & 67.5                       \\  
Pythia \cite{Jiang2018PythiaVT}    & $\--$                 & $\--$                    & 66.3                    & 70.0                       & $\--$                      & $\--$                      \\ 
DFAF \cite{gao2019DFAF}            & $\--$                 & $\--$                    & 66.2                    & 70.2                       & $\--$                      & $\--$                      \\
fPMC \cite{hu2018learningfPMC}     & $\--$                 & $\--$                    & 61.7                    & 63.9                       & $\--$                      & 66.0                       \\ 
STL \cite{wang2018structuredSTL}   & $\--$                 & $\--$                    & $\--$                   & $\--$                      & 67.5                       & 68.2                       \\ 
CTI \cite{do2019cti}               & 61.7                  & 54.9                     & 66.0                    & 70.1                       & 67.0                       & 69.3                       \\ 
MCAN \cite{yu2019mcan}             & $\--$                 & 57.4                     & 67.2                    & 70.6                       & $\--$                      & $\--$                      \\\hline
MuRel \cite{cadene2019murel}       & $\--$                 & $\--$                    & 65.1                    & 68.0                       & $\--$                      & $\--$                      \\ 
ReGAT \cite{li2019regat}           & $\--$                 & $\--$                    & 67.2                    & 70.3                       & $\--$                      & $\--$                      \\  
MMN \cite{chen2021meta}            & $\--$                 & 60.4                     & $\--$                   & $\--$                      & $\--$                      & $\--$                      \\ 
NMS \cite{Hudson2019LearningBA}    & $\--$                 & 63.2                     & $\--$                   & $\--$                      & $\--$                      & $\--$                      \\ 
HAN \cite{kim2020hypergraph}       & $\--$                 & 69.5                     & 65.5                    & 69.1                       & $\--$                      & $\--$                      \\ \hline
LXMERT \cite{tan2019lxmert}        & 59.8                  & 60.0                     & $\--$                   & 72.4                       & $\--$                      & $\--$                      \\ 
OSCAR \cite{li2020oscar}           & $\--$                 & 61.6                     & $\--$                   & 73.6                       & $\--$                      & $\--$                      \\

UNITER-base \cite{chen2020uniter}           & $\--$                 & $\--$                    & $\--$                   & 72.7                      & $\--$                      & $\--$   \\

UNITER-large \cite{chen2020uniter}           & $\--$                 & $\--$                    & $\--$                   & \textbf{73.8}                      & $\--$                      & $\--$                      \\ \hline
\textbf{CFR (ours)}                                 & \textbf{73.6}         & \textbf{72.1}            & \textbf{69.7}           & 72.5              & \textbf{69.8}              & \textbf{71.9}              \\
\hline
\end{tabular}
}
\end{center}
% % \vspace{-0.5cm}
\caption{The accuracy of our method and other approaches on three VQA datasets.
}
\label{tab:sota}
\end{table}

\begin{figure*}[!ht]
  \centering
    \subfigure[]{\includegraphics[width=0.31\linewidth, height=0.47\linewidth]{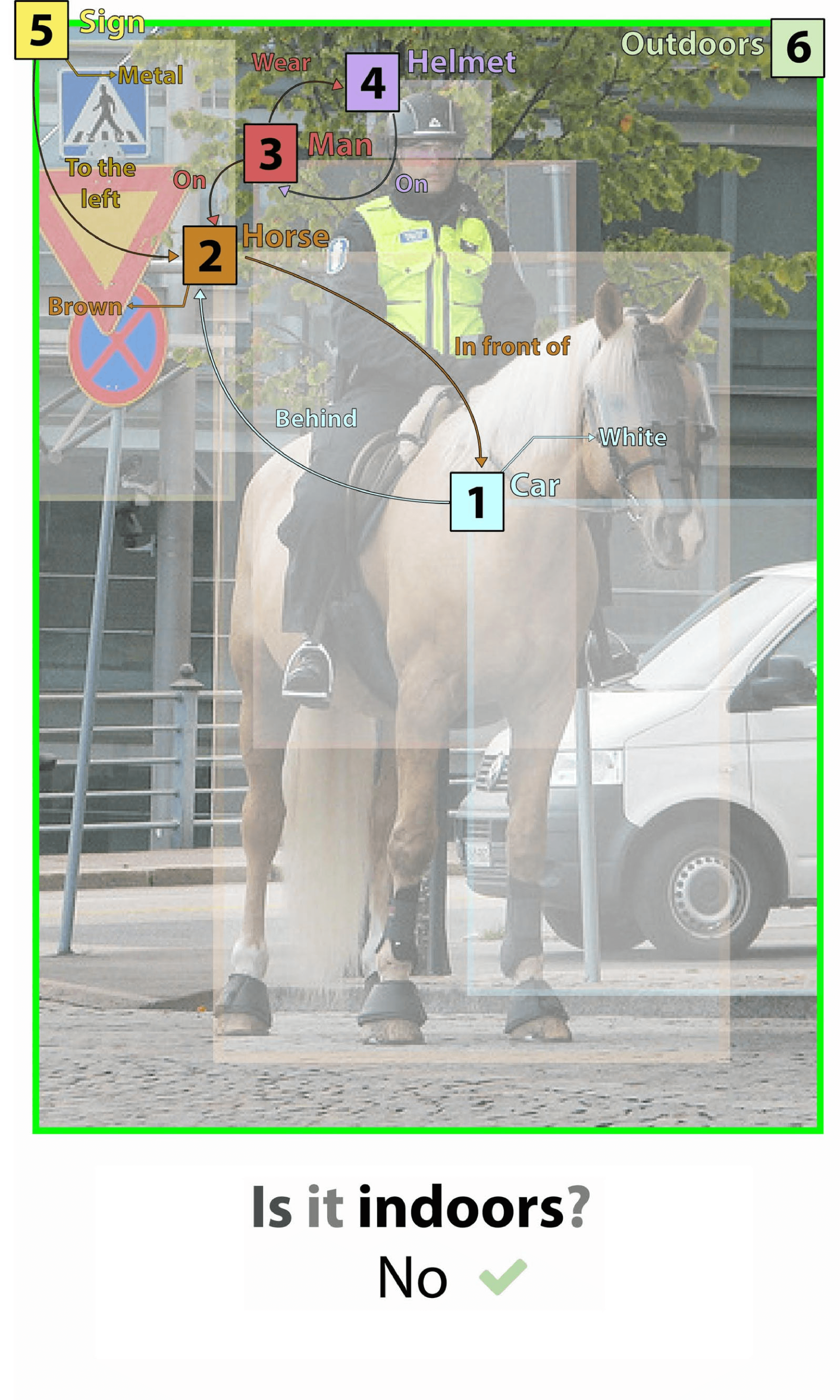}}
    \subfigure[]{\includegraphics[width=0.31\linewidth, height=0.47\linewidth]{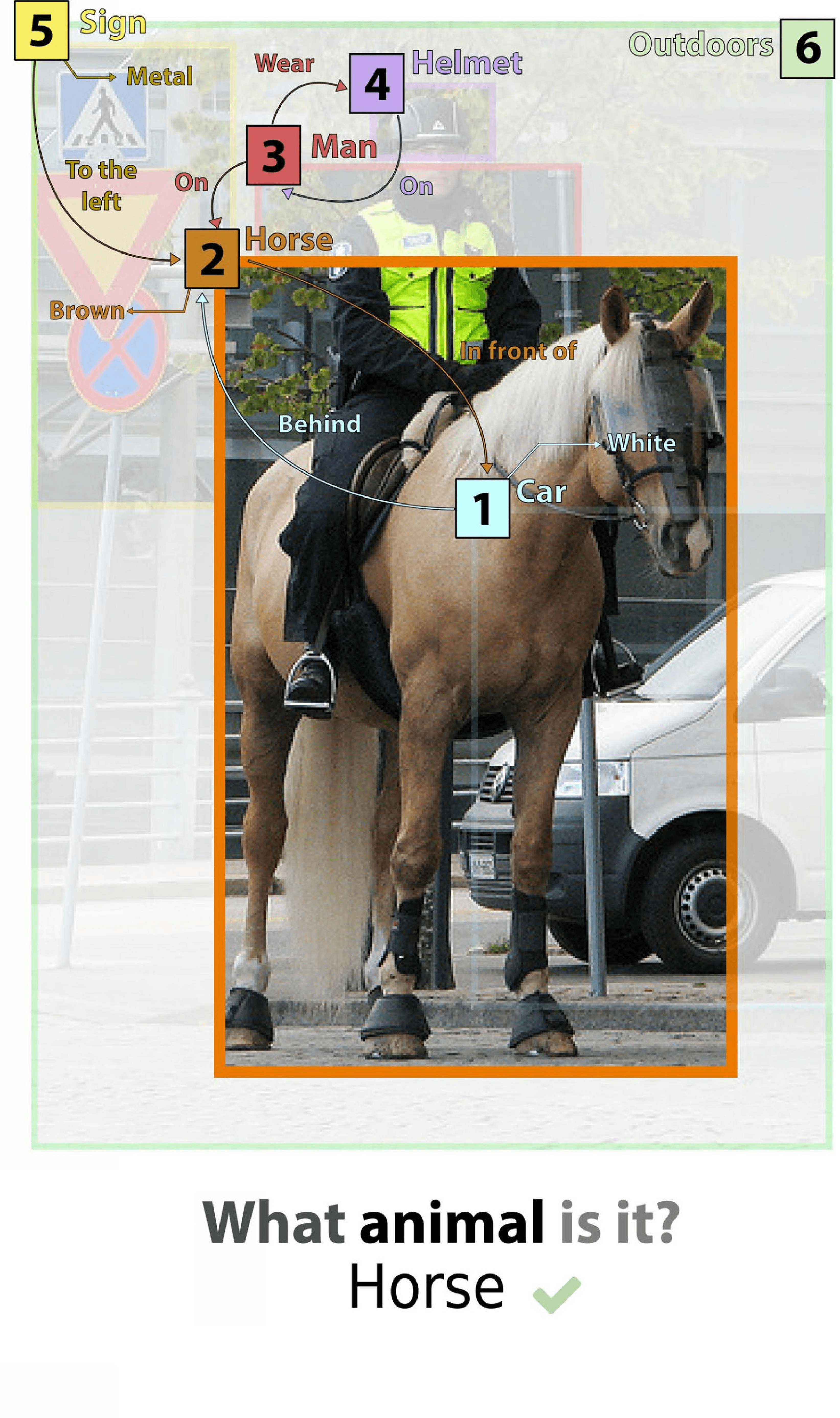}}
    \subfigure[]{\includegraphics[width=0.31\linewidth, height=0.47\linewidth]{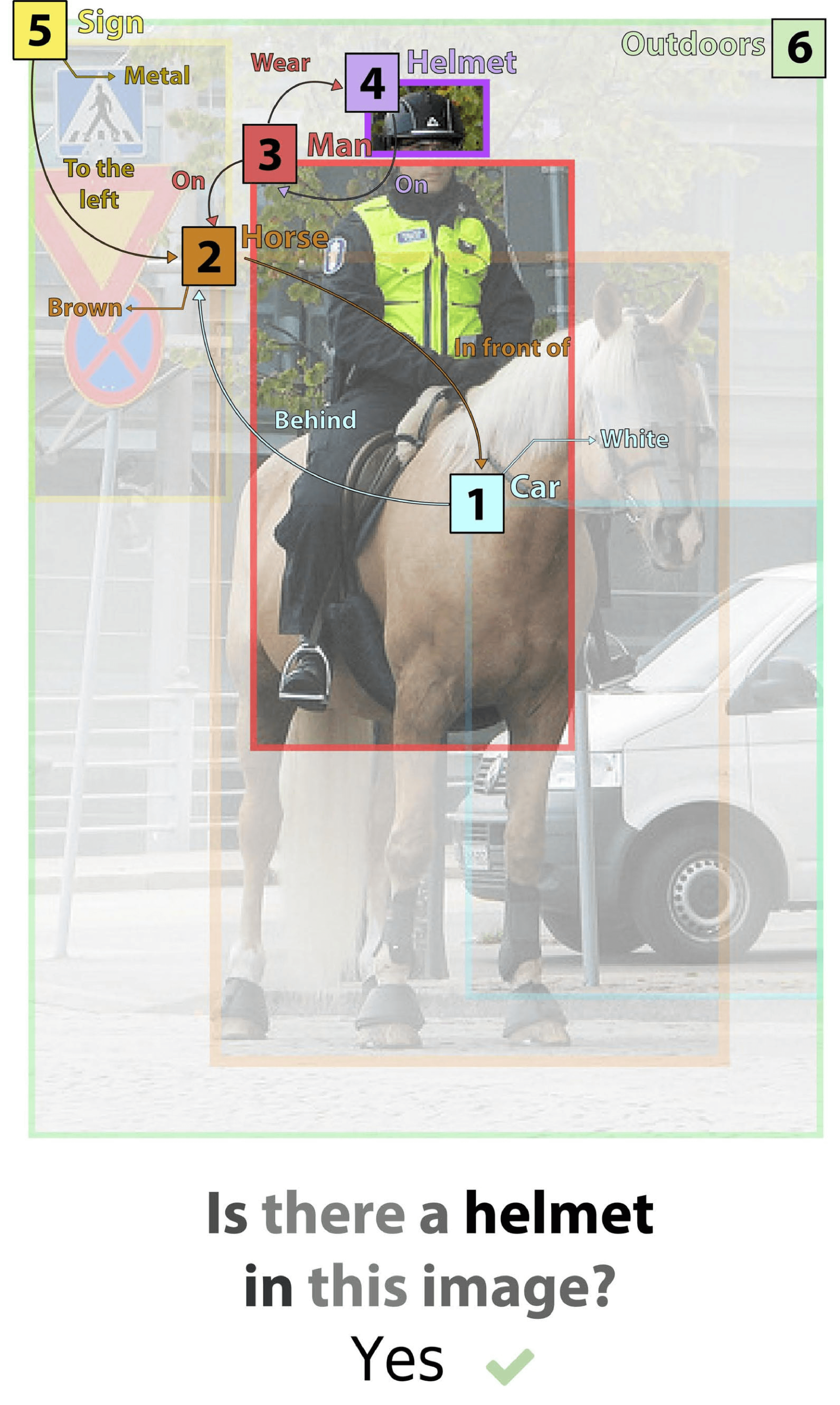}}
    \subfigure[]{\includegraphics[width=0.31\linewidth, height=0.47\linewidth]{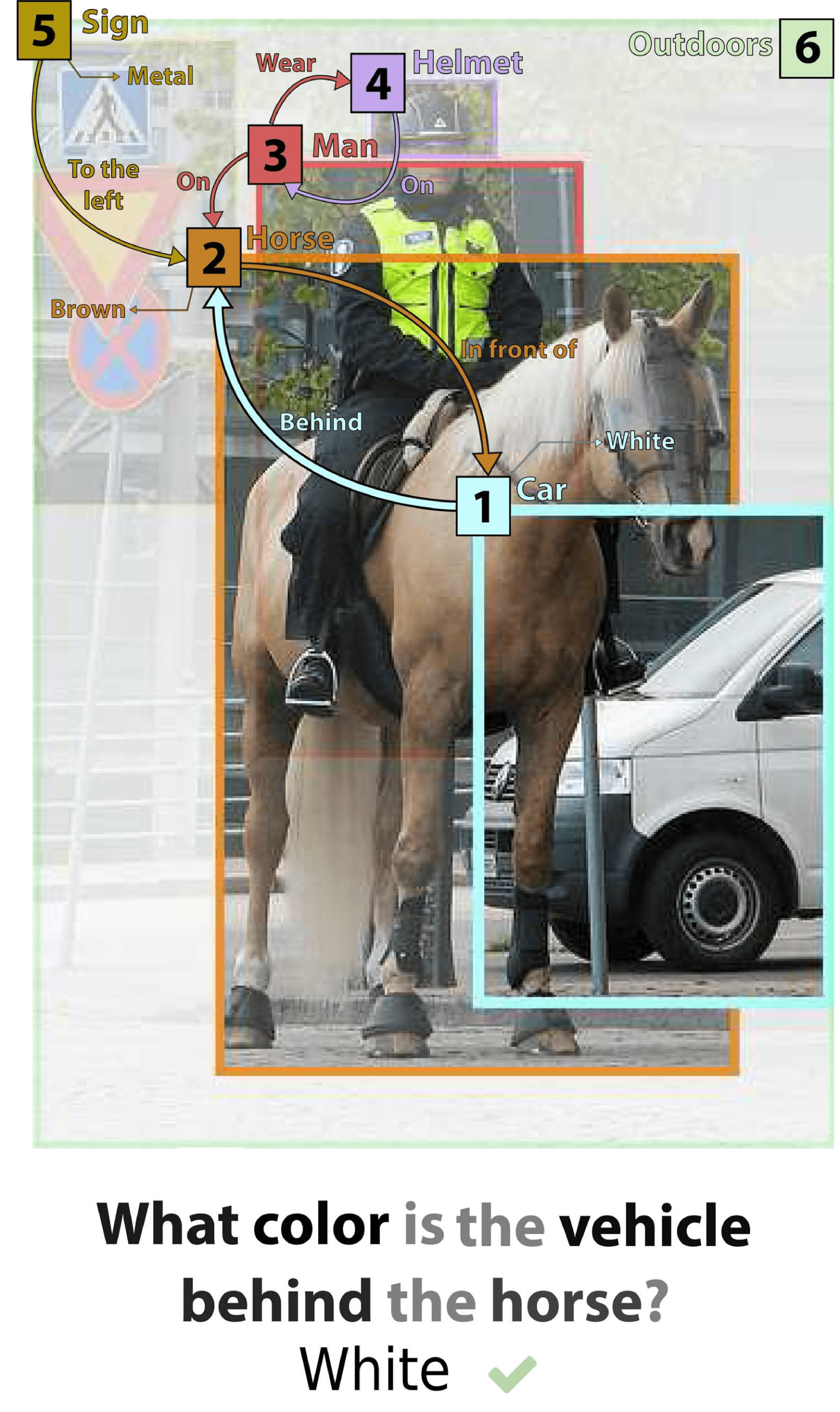}}
    \subfigure[]{\includegraphics[width=0.31\linewidth, height=0.47\linewidth]{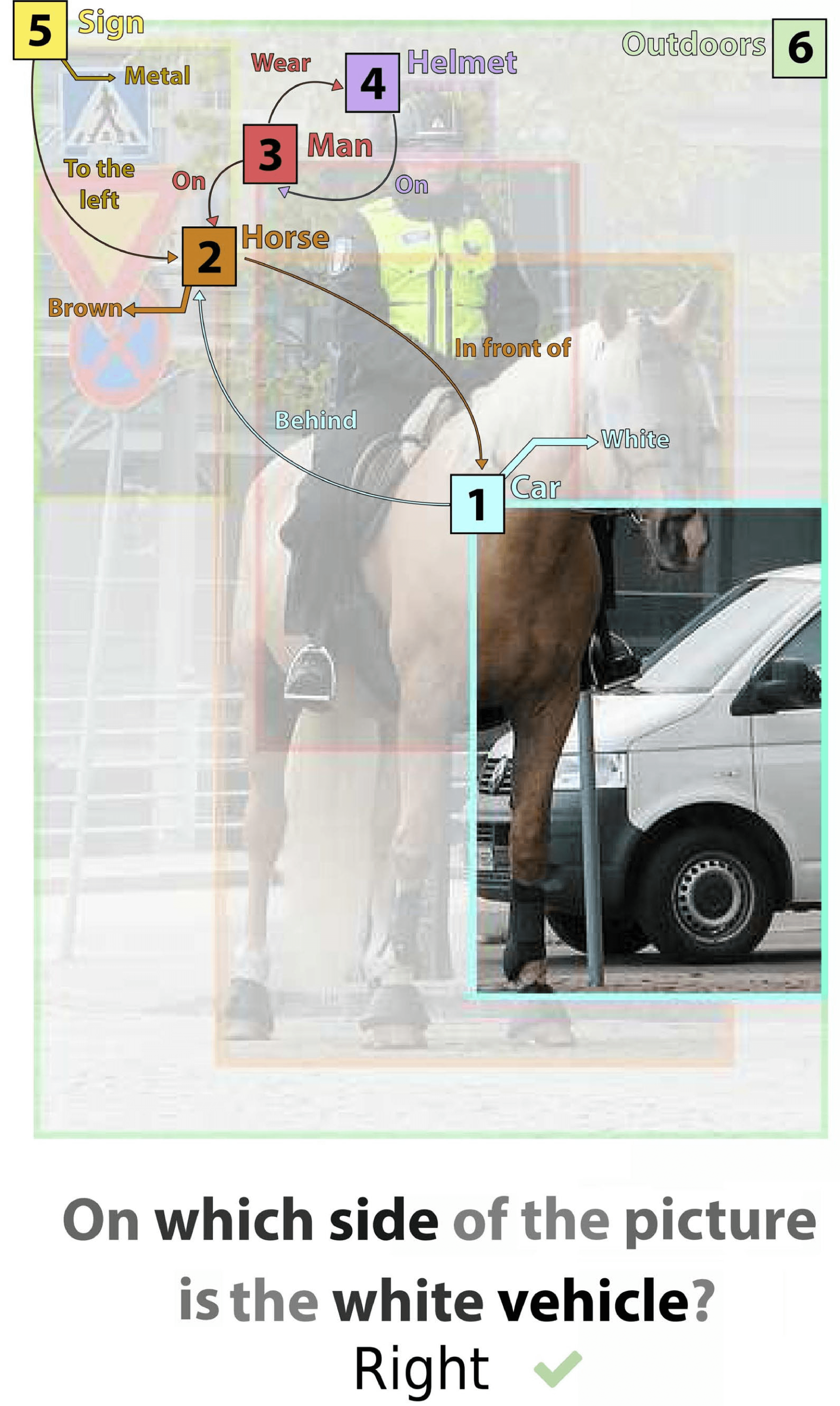}}
    \subfigure[]{\includegraphics[width=0.31\linewidth, height=0.47\linewidth]{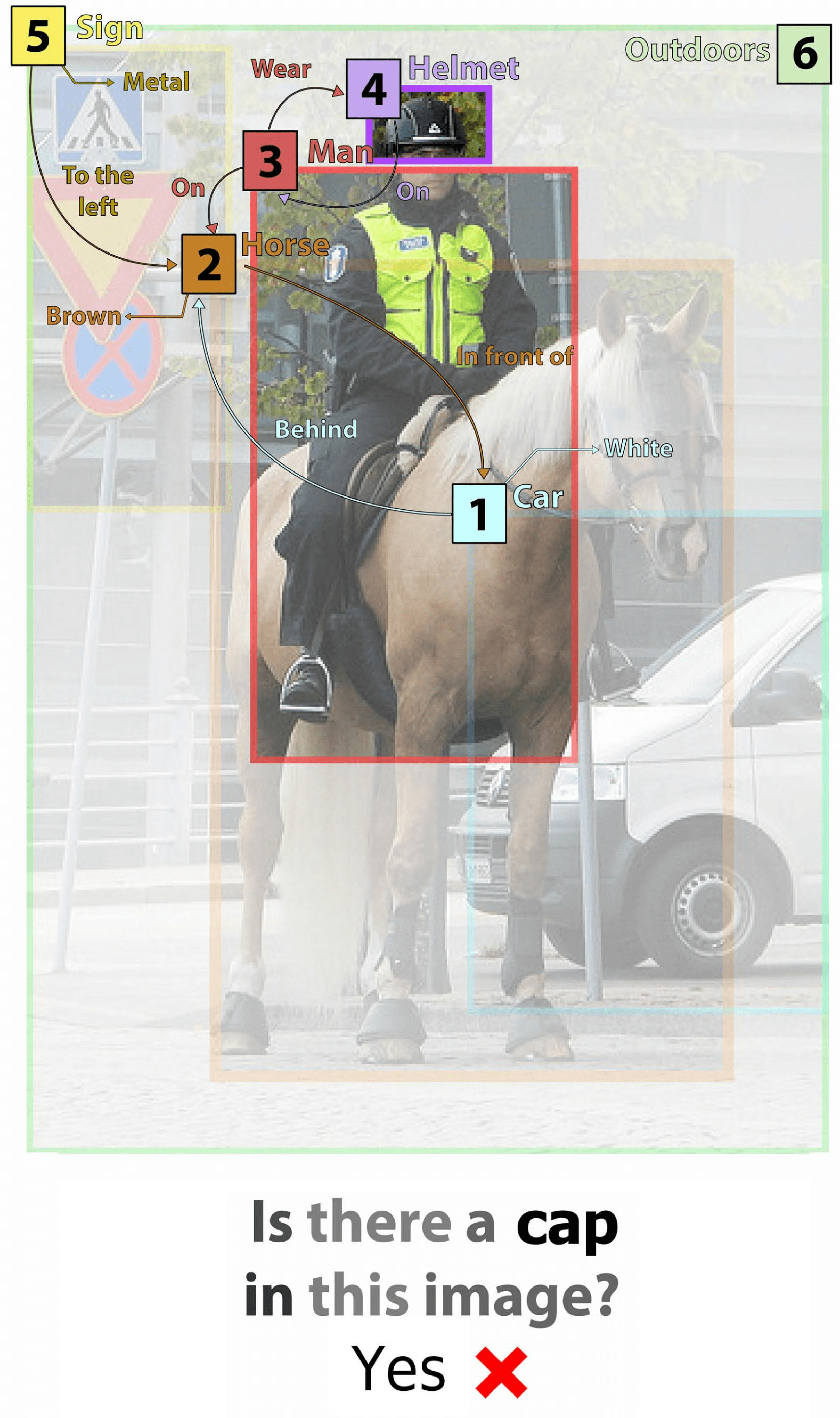}}
    % \vspace{0.1ex}
 \caption{Visualization of the explicit contribution of RoIs and predicates in both input image and question. The \cmark $ $ and  \xmark $ $ symbols indicate the correct and the wrong answers, respectively. The arrow indicates the attribute or relation from the attribute classification or relation classification step in our Image Embedding module.}
 \label{fig:example}
\end{figure*}

\begin{table*}[!ht]
\begin{center}
\small
\setlength{\tabcolsep}{0.17em} % for the horizontal padding
{\renewcommand{\arraystretch}{1.2}% for the vertical padding
\begin{tabular}{|c|c|c|c|c|c|c|c|c|}
\hline
\multicolumn{2}{|c|}{\multirow{3}{*}{\textbf{\begin{tabular}[c]{@{}c@{}} Methods\end{tabular}}}}                                                                          & \multicolumn{3}{c|}{\textbf{Language Modality}}                                                                                                         & \multicolumn{3}{c|}{\textbf{Vision Modality}}                                                                                                           & \multirow{3}{*}{\textbf{\begin{tabular}[c]{@{}c@{}}Acc \\ (\%)\end{tabular}}} \\ \cline{3-8}
\multicolumn{2}{|c|}{}                                                                                                                                                              & \textit{\begin{tabular}[c]{@{}c@{}}Question\\ Features\end{tabular}}                     & \textit{\begin{tabular}[c]{@{}c@{}}Predicates\end{tabular}}                   & \textit{\begin{tabular}[c]{@{}c@{}}Filtered\\ Info\end{tabular}} & \textit{\begin{tabular}[c]{@{}c@{}}Image\\ Features\end{tabular}}                        & \textit{\begin{tabular}[c]{@{}c@{}}Predicates\end{tabular}}                   & \textit{\begin{tabular}[c]{@{}c@{}}Filtered\\ Info\end{tabular}} &                                                                               \\ \hline
%\multicolumn{2}{|c|}{BAN \cite{Kim2018BilinearAN}}                                                                                                                                                  & \cmark\ &                                       &                                                                         & \cmark\ &                                       &                                                                         & 61.5                                                                          \\ \hline
\multirow{3}{*}{\textbf{\begin{tabular}[c]{@{}c@{}}Multimodal\\ Learning\end{tabular}}} & \textit{\begin{tabular}[c]{@{}c@{}}Coars grained\end{tabular}}                & \cmark\ &                                       &                                                                         & \cmark\ &                                       &                                                                         & 62.6                                                                    \\ \cline{2-9} 
                                                                                        & \multirow{2}{*}{\textit{\begin{tabular}[c]{@{}c@{}}Fine grained\end{tabular}}} & \cmark\                                       & \cmark\ &                                                                         & \cmark\                                       & \cmark\ &                                                                         & 67.2 (+4.6)                                                                   \\ \cline{3-9} 
                                                                                        &                                                                                           &\cmark\                                       &                                       & \cmark\                                   &  \cmark\                                     &                                       & \cmark\                                   & 69.5 (+6.9)                                                                   \\ \hline
\multicolumn{2}{|c|}{\textbf{\begin{tabular}[c]{@{}c@{}}Semantic Reasoning\end{tabular}}}                                                                                & \cmark\ &    \cmark\                                      &\cmark\                                 & \cmark\ &  \cmark\                                     & \cmark\                                   & 73.6 (+11.0)                                                                  \\ \hline
\end{tabular}
}
\end{center}
\caption{The contribution of each module in our CFR framework. 
}
\label{tab:abl_module}
\end{table*}
\subsection{Module Contribution}
\label{subsec:abl}

\subsection{Results}
\label{subsec:sota}
Table~\ref{tab:sota} summarizes our results compared with different recent methods in the VQA task. In the GQA dataset, our proposed method outperforms the recent approach HAN~\cite{kim2020hypergraph} on the test-dev set by $+2.6\%$. Regarding the multiple-choice Visual7W dataset, our method outperforms the work CTI \cite{do2019cti} by $2.8 \%$ in the validation set and $2.6 \%$ in the test set, respectively. The results show that our CFR can deal with compositional reasoning questions through the selected information from both coarse-grained learning and fine-grained learning. It is worth noting that our CFR achieves new state-of-the-art results in GQA and Visual7W datasets.

It is more challenging for our method to improve the result in the VQA2.0 dataset. While our CFR still outperforms the recent reasoning work ReGAT \cite{li2019regat} by $2.5\%$ and $2.2\%$, UNITER-large \cite{chen2020uniter} achieves $1.3\%$ higher than our CFR in the test-dev set. We note that the VQA2.0 dataset has fairly fewer compositional reasoning questions comparing with the GQA dataset \cite{hudson2019gqa}. Thus, it limits the effectiveness of methods that focus on reasoning the question and images, including our CFR. Our method also uses simple modules to extract image and question features, which may not be robust enough comparing with features extracted from complicated modules such as large-scale visual-language models \cite{li2020oscar} \cite{chen2020uniter}.

To evaluate the contribution of each module in our framework, we conduct the following experiment: Given different level of information of language and vision modality (features, predicates, and filtered information of the image/question), we gradually choose different pairs of vision and language modality as the input to predict the answer. The experiment is conducted using the GQA dataset.

Table \ref{tab:abl_module} shows the contribution of each module when different inputs are used. By using only the question and image feature (coarse-grained learning), our framework only achieves $62.6\%$ accuracy. When we combine the question and image features with their corresponding predicates (fine-grained learning), the accuracy increase to $67.2 \%$. This result indicates the effectiveness of predicates. By applying the filtered information of both question and image, the performance of fine-grained learning increases to $69.5\%$. This result shows that by reducing the negative influence of noisy information, the prediction accuracy can be improved. To effectively leverage all coarse-grained and fine-grained information, the Semantic Module is integrated into the framework and achieves $73.6 \%$ accuracy. This result validates the potency of Semantic Reasoning in selecting information for answering the complicated question. Overall, our introduced framework outperforms the baseline coarse-grained learning method by a large margin, i.e., $+11.0 \%$ accuracy.

\subsection{Visualization} 

% %%%% TEMPO DISABLE
% %%%%%%%%%%%%%%%%%%%%%%%%%%%%%%%%%%%%%%%

Figure \ref{fig:sample_Adaptation} illustrates the comparison between using Coarse-grained learning, Fine-grained learning, and Semantic Reasoning when we visualize the confidence score of the top 5 output answers. From this figure, we notice that if the Coarse-grained or Fine-grained learning are used separately, the output answer may not be correct, and there is usually an ambiguity in the top two predicted answers. However, when we apply our whole Coarse-to-Fine Reasoning framework, the network predicts both answers correctly, and also there is no ambiguity between the top predicted answer and the second predicted answer. These results show that our Coarse-to-Fine Reasoning framework successfully encodes both the features and predicates from the image and question in a coarse-to-fine manner, hence consequently improves the prediction results.

Figure \ref{fig:example} illustrates the explicit contribution of RoIs and predicates in both input image and question when our framework answers different compositional questions. Note that the transparency level of each RoI/word indicates the importance of each information. The RoIs and predicates with no opacity are crucial instances for answering the corresponding question. The visualizations in samples $\rm (a,b,c,d,e)$ indicate the effectiveness of our CFR framework in reasoning the correct answers from the inference process. The sample in $\rm (f)$ demonstrates the case when our CFR predicts the wrong answer. The incorrect prediction may come from the limitation of extractors, i.e., the extracted features are not robust enough (e.g., ``cap" and ``helmet" in our false example). Figure \ref{fig:example} also shows that our CRF framework not only can increase the accuracy of the VQA task but also provides an explainable way to understand the prediction results.

\section{Conclusion}
\label{sec:con}
We have introduced a new simple, yet effective Coarse-to-Fine Reasoning (CFR) framework for the VQA task. Our CRF framework first extracts the features and predicates of both question and image. Then we propose a new reasoning module to map the key information in the question to the visual clues in the image in a coarse-to-fine manner. The intensive experiments on GQA, VQA2.0, and Visual7W datasets show that our framework achieves competitive results comparing with recent approaches. Our source code and trained models will be released for reproducibility and further study.

%%%%%%%%% REFERENCES
{\small
\bibliographystyle{ieee_fullname}
\bibliography{egbib}
}

\end{document}